\documentclass{article}

\usepackage[preprint, nonatbib]{neurips_2019}

\usepackage[utf8]{inputenc} 
\usepackage[T1]{fontenc}    
\usepackage{hyperref}       
\usepackage{url}            
\usepackage{booktabs}       
\usepackage{amsfonts}       
\usepackage{nicefrac}       
\usepackage{microtype}      
\usepackage{amsmath}        
\usepackage{algorithm,algorithmic}
\usepackage{color}
\usepackage{graphicx}

\DeclareRobustCommand{\bm}[1]{\mathbf{\boldsymbol{#1}}}
\DeclareRobustCommand{\im}[0]{{{\rm i}\mkern1mu}}

\newcommand{\Diag}[0]{\mathop{\rm Di}}

\title{Replicated Vector Approximate Message Passing For Resampling Problem}

\author{%
  Takashi Takahashi and Yoshiyuki Kabashima\\
  Department of Mathematical and Computing Science\\
  Tokyo Institute of Technology\\
  2-12-1, Ookayama, Meguro-ku, Tokyo, Japan \\
  \texttt{\{takahashi, kaba\}@sp.dis.titech.ac.jp} 
}

\begin{document}

\maketitle

\begin{abstract}
    Resampling techniques are widely used in statistical inference and ensemble learning, in which estimators' statistical properties are essential.
    However, existing methods are computationally demanding, because repetitions of estimation/learning via numerical optimization/integral for each resampled data are required. 
    In this study, we introduce a computationally efficient method to resolve such problem: \emph{replicated vector approximate message passing}.
    This is based on a combination of the replica method of statistical physics and an accurate approximate inference algorithm, namely the \emph{vector approximate message passing} of information theory.
    The method provides tractable densities without repeating estimation/learning, and the densities approximately offer an arbitrary degree of the estimators' moment in practical time.
    In the experiment, we apply the proposed method to the stability selection method, which is commonly used in variable selection problems.
    The numerical results show its fast convergence and high approximation accuracy for problems involving both synthetic and real-world datasets.

\end{abstract}

\section{Introduction}
    A widely accepted strategy in statistics and machine learning involves leveraging statistical properties of estimators concerning obtained datasets and model hyperparameters, to improve the quality of inference and learning.
    Examples of such techniques range from variable selection methods in high-dimensional statistics \cite{meinshausen2010stability, dezeure2015high, dezeure2017high} to bagging techniques in machine learning \cite{breiman1996bagging, wallace2011class}. 
    Modern statistical models rarely exhibit a closed form of estimators; hence, most procedures are computationally performed due to necessity.
    The procedures typically consist of Monte-Carlo (MC) resampling of datasets/hyperparameters and repetitions of estimation/learning for pseudo-sample obtained by MC resampling via mathematical optimization/integral.
    
    The use of the techniques mentioned above causes two problems.
    The first corresponds to the computational cost due to the re-estimation/re-training for each MC sample.
    Resampling techniques for modern statistical methods require many samples; hence, it can entail substantial computational time.
    The second is concerned with a theoretical issue.
    Generally, it is difficult to characterize the distribution of estimators for the resampled data analytically.
    This difficulty prevents gaining useful insights from quantitative theoretical analysis.
    
    In this study, we address the former problem of heavy computational cost.
    We introduce a computationally efficient approximate inference scheme.
    The proposed approach is based on the replica method of statistical physics \cite{mezard1987spin} and \emph{vector approximate message passing} (VAMP) of information theory \cite{rangan2017vector} that corresponds to a systematic and highly accurate approximate inference algorithm.
    The combination of these two techniques gives a computationally efficient approximate inference method to offer the estimators' distribution without repeated estimation.
    We apply the proposed method to stability selection \cite{meinshausen2010stability} that is widely used in variable selection problems.
    The numerical results indicate that our method exhibits fast convergence and achieves accurate estimates for both synthetic and real-world data.
    
    \subsection{Related work}

    \cite{malzahn2003approximate, NIPS2002_2185} initially introduced a fairly general strategy for the resampling problem based on the replica method and sophisticated variational method, and demonstrated its potential usefulness.
    However, there is less progress in this direction due to the unmanageable convergence property and intricate derivation.
    Because their variational methods were initially developed to analyze the theoretical properties of probabilistic models, the construction of efficient algorithms to obtain the approximate densities itself was lacking at that time.
    
    There was significant progress on the aforementioned algorithmic problem in studies related to information theory due to the discovery of \emph{approximate message passing} (AMP)  algorithms.
    Specifically, AMP was initially introduced as a computationally efficient iterative signal recovery algorithm with a rigorous guarantee of convergence in the context of CDMA multiuser detection and compressed sensing \cite{kabashima2003cdma, donoho2009message}.
    \cite{kabashima2003cdma, fletcher2016expectation, rangan2017vector} showed that AMP and its generalizations share the same fixed points as the iterative formulae of variational methods including the adaptive Thouless-Anderson-Palmer (TAP) method \cite{opper2001adaptive, opper2001tractable} and Expectation Consistent (EC) approximate inference \cite{opper2005expectation}.
    \cite{meng2015expectation, rangan2017vector} discussed relations between AMP algorithms and expectation propagation (EP) \cite{minka2001expectation}, and provided a systematic derivation.
    
    Recently, \cite{obuchi2018semi} derived an AMP-based approximate resampling algorithm and described its convergence dynamics.
    However, its application is limited to a rather restricted class of problems.

\section{Resampling problem}
    \label{sec:resampling problem}
    We assume that there is a dataset $D = \{z_\mu\}_{\mu=1}^M$, and introduce an associated likelihood of form
    \begin{equation}
        p(D\mid f) \propto e^{-\sum_{\mu=1}^M l(z_\mu; f)},
    \end{equation}
    where the negative log-likelihood $l(z_\mu; f)$ for the data point $z_\mu$ is characterized via a function $f$ that represents a model output.
    Then, with respect to an appropriate prior distribution $p_0(f;\Theta)$ parametrized by a hyperparameter $\Theta$, the posterior distribution is defined as
    \begin{equation}
        p^{(\beta)}(f\mid D) = \frac{1}{Z^{(\beta)}(D, \Theta)}p(D\mid f)^\beta p_0(f; \Theta)^\beta, \label{eq:posterior}
    \end{equation}
    where $\beta>0$ is termed as the inverse temperature, and $Z^{(\beta)}(D, \Theta)$ denotes the normalization constant called the partition function: $Z^{(\beta)}(D, \Theta) = \int p(D\mid f)^\beta p_0(f; \Theta)^\beta df$.
    The case in which $\beta=1$ corresponds to strict Bayes inference.
    The limit $\beta\to\infty$ corresponds to maximum a posteriori (MAP) estimation because the distribution (\ref{eq:posterior}) concentrates on the global maxima of the original posterior distribution \cite{mezard2009information}. 
    Although $l(z;f)$ and $p_0$ are referred to as log-likelihood and prior distribution, respectively, it is generally not necessary to provide a strict probabilistic interpretation.
    In case when we explicitly model the function $f$ by parameters, the distribution of $f$ corresponds to its parameters' distribution.
    We focus on a statistical estimator obtained as a posterior average of statistics $A(f)$:
    \begin{equation}
        \hat{A}^{(\beta)}(D,\Theta) = \mathbb{E}_{f}\left[A(f); p^{(\beta)}(f\mid D;\Theta)\right]. \label{eq:estimator}
    \end{equation}
    The notation $\mathbb{E}_x[...; p(x)]$ denotes the average for a random variable $x$ distributed over a probability density $p$.
    We omit the argument $p$ when there is no risk of confusion.
    The purpose of the resampling problem involves evaluating the statistical property of $\hat{A}^{(\beta)}(D, \Theta)$ when $D, \Theta$ are distributed.

\section{Replica method for resampling problem}
    In this section, we describe the strategy of approximate inference to avoid repeated estimation/learning based on the replica trick of statistical physics.
    Using the definition of the estimator $(\ref{eq:estimator})$ and the posterior $(\ref{eq:posterior})$, the estimator's moment is expressed as 
    \begin{equation}
        \mathbb{E}_{D,\Theta}\left[\left\{\hat{A}^{(\beta)}(D, \Theta)\right\}^r\right] = \int \prod_{a=1}^r A(f^{(a)}) \mathbb{E}_{D,\Theta}\left[\prod_{a=1}^r\frac{p(D\mid f^{(a)})^{\beta}p_0(f^{(a)};\Theta)^{\beta}}{Z^{(\beta)}(D,\Theta)}\right]d^r f, \label{eq:r-replicas}
    \end{equation}
    where we introduce the notation $d^k f = df^{(1)}df^{(2)}...df^{(k)}, k\in\mathbb{N}$.
    It is difficult to evaluate analytically due to the presence of the partition function that depends on $D$ and $\Theta$ in the denominator, which is the origin of repeated numerical estimation/learning.
    The replica trick \cite{mezard1987spin} bypasses the problem via an identity $Z^{-r} = \lim_{n\to0}Z^{n-r}$.
    Using this identity, (\ref{eq:r-replicas}) is re-expressed as 
    \begin{equation}
        \mathbb{E}_{D,\Theta}\left[\left\{\hat{A}^{(\beta)}(D,\Theta)\right\}^r\right] = \lim_{n\to 0}\mathcal{A}_n^{(\beta)},
        \label{eq:replica_identity}
    \end{equation}
    where 
    \begin{equation}
        \mathcal{A}_n^{(\beta)} = \int \prod_{a=1}^r A(f^{(a)}) \mathbb{E}_{D,\Theta}\left[\left\{Z^{(\beta)}(D,\Theta)\right\}^{n-r}\prod_{a=1}^r p(D\mid f^{(a)})^\beta p_0(f^{(a)};\Theta)^\beta\right]d^rf.
    \end{equation}
    The advantage of the formula is that for integer $n \ge r$, the negative power of the partition function $\{Z^{(\beta)}(D, \Theta)\}^{-r}$ in (\ref{eq:r-replicas}) is eliminated by $n$ replicas of variables using the integral form of $Z$:
    \begin{equation}
        \mathcal{A}_n^{(\beta)} = \Xi_n \int \prod_{a=1}^r A(f^{(a)}) \frac{1}{\Xi_n}\mathbb{E}_{D,\Theta}\left[\prod_{a=1}^n p(D\mid f^{(a)})^{\beta} p_0(f^{(a)} ; \Theta)^{\beta}\right] d^n f,
    \end{equation}
    which is amenable to analytical approximation techniques.
    We introduce the normalization constant $\Xi_n=\int \mathbb{E}_{D,\Theta}\left[\prod_{a=1}^n p(D\mid f^{(a)})^{\beta}p_0(f^{(a)};\Theta)^\beta\right]d^nf$ to normalize the measure of $f$.
    Because by construction $\lim_{n\to0}\Xi_n=1$, we omit $\Xi_n$ in the following.
    Given expression of (\ref{eq:replica_identity}), we calculate $\mathcal{A}_n^{(\beta)}$ as if $n$ were an integer.
    After obtaining a sufficiently manageable expression with respect to $n$, we extrapolate $n$ to $\mathbb{R}$ and take the limit $n\to0$.
    In the following, we use the notation $\bm{f}=(f^{(1)}, f^{(2)}, ..., f^{(n)})$ and the symbol $\doteq$ to denote equality up to a normalization constant.
    With these notations, we call the probability density function
    \begin{equation}
        \tilde{p}^{(\beta)}(\bm{f}) \doteq \mathbb{E}_{D,\Theta}\left[\prod_{a=1}^n p(D\mid f^{(a)})^{\beta}p_0(f^{(a)};\Theta)^\beta\right], \label{eq:replicated_system}
    \end{equation}
    the \emph{replicaed system}.
    The replicated system $\tilde{p}^{(\beta)}(\bm{f})$ is intrinsically invariant under all permutations of $\{f^{(1)}, f^{(2)},...,f^{(n)}\}$.
    The property is termed as the {\em replica symmetry} (RS). 
    Then, de Finetti's representation theorem \cite{hewitt1955symmetric} guarantees that (\ref{eq:replicated_system}) is re-expressed as  
    \begin{equation}
        \tilde{p}^{(\beta)}(\bm{f}) = \int \prod_{a=1}^n p^{(\beta)}(f^{(a)} \mid \xi)  p^{(\beta)}(\xi)d\xi, \label{eq:de-finetti}
    \end{equation}
    where $\xi$ is some random variable which directly reflects the effect of $D$ and $\Theta$.
    This expression indicates that the estimator's moment is reduced to a considerably simple form:
    \begin{equation}
        \mathbb{E}_{D,\Theta}\left[\left\{\hat{A}^{(\beta)}(D,\Theta) \right\}^r\right] = \int \left\{\mathbb{E}_{f}\left[ A(f) ; p^{(\beta)}(f\mid\xi)\right]\right\}^r p^{(\beta)}(\xi)d\xi. \label{eq:rs-form-of-estimator}
    \end{equation}
    Thus, we can obtain an arbitrary degree of the moment without repetition of estimation/learning, by obtaining tractable densities for $p^{(\beta)}(f\mid\xi)$ and $p^{(\beta)}(\xi)$ under an appropriate approximation.
    
\section{Replicated vector approximate message passing}
    \label{sec:rVAMP}
    
    In this section, we introduce a concrete algorithm to obtain approximate densities for $p^{(\beta)}(f\mid\xi)$ and $p^{(\beta)}(\xi)$.
    The derivation is based on a variable augmentation via Fourier transform representation of the delta function, and a message passing form of EP on a factor graph wherein variable nodes represent sets of replicated vectors.
    
    \begin{figure}[t]
      \centering
      \includegraphics[width= \columnwidth]{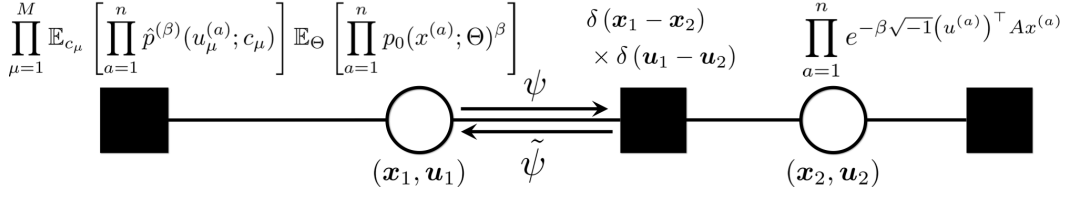}
      \caption{Factor graph representation used for the derivation of rVAMP. Circles represent variable nodes and squares represent factor nodes.
      The message from a factor node to a variable node is denoted by $\psi$, and the message in the opposite direction is denoted by $\tilde{\psi}$.\vspace{-2pt}}
      \label{fig:factor_graph}
    \end{figure}
    
    \subsection{Problem setup}
    In the following, we consider the specific case that the function $f$ is modeled via a single layer model.
    In this case, the value of $f$ for each data point $z_\mu$ is given as $a_\mu^{\top} x$ where $a_\mu$ denotes a feature vector known in advance, and the distribution of function $f$ is replaced with that of parameter $x\in\mathbb{R}^N$.
    Furthermore, it is necessary to specify the distribution of $D$ and $\Theta$.
    Typically, the distribution of the hyperparameter $p(\Theta)$ is explicitly given based on each resampling method.
    Conversely, the distribution of the input data $q(D) = \prod_{\mu}q(z_\mu)$ is unknown.
    Thus, we replace it with \emph{bootstrap distribution} \cite{efron1994introduction} that approximates the distribution of observed data with its empirical distribution: $q(D) \simeq \hat{q}(D) = \prod_\mu \hat{q}(z_\mu)$, where $\hat{q}(z) = M^{-1}\sum_{\nu}\delta(z-z_\nu)$.
    Specifically, the empirical distribution $\hat{q}$ is introduced as an unbiased estimator of the true data distribution $q(D)$.
    Subsequently, a resampled dataset $D^*$ of size $M_{\rm B}$ from $\hat{q}$ is represented by a \emph{occupation vector} $c=(c_1,...,c_M)\in\{0, 1,...,M_B\}^{M}$ with $\sum_{\mu=1}^{M}c_\mu = M_{\rm B}$.
    $c_\mu$ denotes the number of times that the data point $z_\mu$ appears in the set $D^*$.
    The strict distribution of $c$ is multinomial.
    However, for large $M$, we can replace it with a product of Poisson distribution with mean $\tau \equiv M_{\rm B}/M$ \cite{NIPS2002_2185}: $\hat{q}(D^*) = \hat{q}(c) \simeq \prod_{\mu = 1}^{M}e^{-\tau}\tau^{c_\mu}/c_\mu!$.
    
    \subsection{Augmented replicated system}
    With respect to the problem setup described above, the replicated system is expressed as follows:
    \begin{equation}
        \tilde{p}^{(\beta)}(\bm{x}) \doteq \prod_{\mu=1}^M \mathbb{E}_{c_\mu}\left[\prod_{a=1}^n e^{- \beta c_\mu l\left(z_\mu; a_\mu^\top x^{(a)}\right)}\right]
        \mathbb{E}_{\Theta}\left[\prod_{a=1}^n p_0(x^{(a)};\Theta)^\beta \right].  \label{eq:replicated_system_specific}
    \end{equation}
    The goal is to obtain a tractable density of the replicated system.
    To this aim, we re-express (\ref{eq:replicated_system_specific}) via the Fourier transform representation of the delta function $\delta(s)=\int e^{- \beta \im us}\beta du/(2\pi)$ for $ s\in\mathbb{R}$:
    \begin{align}
        \label{eq:augmented_replica_system}
        \tilde{p}^{(\beta)}(\bm{x}) 
        \doteq& \int \prod_{\mu=1}^M \mathbb{E}_{c_\mu}\left[\prod_{a=1}^n \hat{p}^{(\beta)}(u_{\mu}^{(a)}; c_\mu)\right] \mathbb{E}_{\Theta}\left[\prod_{a=1}^n p_0(x^{(a)}; \Theta)^\beta\right]
        \prod_{a=1}^n e^{-\im (u^{(a)})^\top A x^{(a)}}
        d^n u, 
    \end{align}
    where $\hat{p}^{(\beta)}(u_\mu^{(a)}; c_\mu) \doteq \int e^{-\beta \left\{ c_\mu l\left(z_\mu ; f_\mu^{(a)}\right) +\im u_\mu^{(a)} f_\mu^{(a)}\right \}}df_\mu^{(a)}$ and $\im\equiv\sqrt{-1}$.
    Thus, the replicated system is expressed as the marginal of the joint distribution $\tilde{p}^{(\beta)}(\bm{x}, \bm{u})$ with original variable $\bm{x}$ and augmented variable $\bm{u}=(u^{(1)}, ...,u^{(n)})\in\mathbb{R}^{M\times n}$. 
    We call $\tilde{p}^{(\beta)}(\bm{x}, \bm{u})$ \emph{augmented replicated system}.

    \subsection{Replicated VAMP}

    Our idea involves approximating the augmented replicated system $\tilde{p}^{(\beta)}(\bm{x}, \bm{u})$ via VAMP and then taking the marginal.
    Because the augmented replicated system can be viewed as a posterior with the Gaussian form of likelihood $\prod_{a=1}^n \exp(-\beta\im (u^{(a)})^\top A x^{(a)})$, the strategy developed in \cite{rangan2017vector} is immediately applicable.
    In order to derive VAMP on the augmented replicated system, we split variables $\bm{x}, \bm{u}$ into two equivalent variables $\bm{x}_1, \bm{u}_1$ and $\bm{x}_2, \bm{u}_2$, to yield an equivalent distribution:
    \begin{align}
        \label{eq:variable_splitting}
        \tilde{p}^{(\beta)}(\bm{x}_1, \bm{u}_1, \bm{x}_2, \bm{u}_2)
        \doteq \left\{\prod_{\mu=1}^M \mathbb{E}_{c_\mu}\left[\prod_{a=1}^n \hat{p}^{(\beta)}(u_{1,\mu}^{(a)}; c_\mu)\right] \mathbb{E}_{\Theta}\left[\prod_{a=1}^n p_0(x_1^{(a)}; \Theta)^\beta\right]\right\}
        \nonumber \\
        \times \delta(\bm{x}_1 - \bm{x}_2)\delta(\bm{u}_1 - \bm{u}_2) \left\{\prod_{a=1}^n e^{-\beta\im \left(u_2^{(a)}\right)^\top A x_2^{(a)}}\right\},
    \end{align}
    The factor graph corresponding to (\ref{eq:variable_splitting}) is shown in Figure \ref{fig:factor_graph}.
    VAMP is derived by applying a message passing form of EP \cite{minka2005divergence} to the factor graph whose variable nodes represent replicated vectors.
    Specifically, $\psi_{i\to \mathcal{F}}(\bm{x}_i, \bm{u}_i)$ and $\tilde{\psi}_{\mathcal{F}\to i}(\bm{x}_i, \bm{u}_i)$ denote the message from variable node $i$ to factor node $\mathcal{F}$ and message in the opposite direction, respectively, and the messages are given as follows:
    \begin{align}
        \label{eq:message_variable_to_factor}
        \psi_{i\to\mathcal{F}}(\bm{x}_i, \bm{u}_i) &\doteq \frac{1}{\tilde{\psi}_{\mathcal{F}\to i}(\bm{x}_i,\bm{u}_i)}{\rm Proj}_{\Phi}\left[\prod_{\mathcal{G}\in \partial i}\tilde{\psi}_{\mathcal{G}\to i}(\bm{x}_i,\bm{u}_i)\right], \\
        \label{eq:message_factor_to_variable}
        \tilde{\psi}_{\mathcal{F}\to i}(\bm{x}_i, \bm{u}_i) &\doteq \frac{1}{\psi_{i\to \mathcal{F}}(\bm{x}_i, \bm{u}_i)}{\rm Proj}_{\Phi}\left[\int\mathcal{F}(x_{\partial \mathcal{F}}, u_{\partial \mathcal{F}}) \prod_{j \in \partial \mathcal{F}}\psi_{j\to \mathcal{F}}(\bm{x}_j, \bm{u}_j) dx_{\setminus i}du_{\setminus i}\right],
    \end{align}
    where $\partial i$ denotes the set that contains all factors involved in variable node $i$, and similarly $\partial{\mathcal{G}}, \partial{\mathcal{F}}$ denotes the set that contains all variables involved in factor node $\mathcal{F}, \mathcal{G}$.
    $x_{\setminus i}, u_{\setminus i}$ denote all the variables except variable $i$.
    Furthermore, the projection operator is defined as ${\rm Proj}_{\Phi}[p] = \arg\min_{q\in\Phi}{\rm KL}[q||p]$ for a density $p$ and a probability density family $\Phi$.
    If we constrain $\Phi$ to be exponential families, the projection leads to moment matching between $q$ and $p$.
    In order to ensure that the message calculation is tractable, we set $\Phi$ as the following form of $(N + M)n$-dimensional multivariate Gaussians:
    \begin{equation}
        \label{eq:gaussian_family}
        \Phi = \left\{
            \begin{array}{c|c}
                \begin{split}
                    \mathcal{N}\left(\bm{x}; \bm{r}_x, \Lambda_x^{-1}\right) \\
                    \times \mathcal{N}\left(\bm{u}; \bm{r}_u, \Lambda_u^{-1}\right)
                \end{split}
                &
                \begin{split}
                    r_{x,i}^{(a)} &= r_{x,i}, \quad 
                    r_{u,\mu}^{(a)} = r_{u, i}, \\
                    \Lambda_{x, ij}^{(ab)} &= \delta_{ij}[\delta_{ab}(\beta\hat{Q}_{x,i} - \beta^2\hat{\chi}_{x,i}) + (1-\delta_{ab})(-\beta^2\hat{\chi}_{x,i})], \\
                    \Lambda_{u, \mu \nu}^{(ab)} &= \delta_{\mu\nu}[\delta_{ab}(\beta\hat{Q}_{u,\mu} + \beta^2\hat{\chi}_{u, \mu}) + (1-\delta_{ab})(\beta^2\hat{\chi}_{u, \mu})], \\
                    i,j &= 1,2,..,N, \; \mu, \nu = 1,2,..,M, \; a,b = 1,2,...,n
                \end{split}
            \end{array}
        \right\},
    \end{equation}
    where $\mathcal{N}(x;\mu, \Sigma)$ denotes a Gaussian density with mean $\mu$ and covariance $\Sigma$.
    Thus, $\Phi$ denotes factorized Gaussian densities that retains an RS form of correlation only between replicas, which corresponds to the \emph{RS ansatz} \cite{mezard1987spin}.
    Employing expression of (\ref{eq:de-finetti}) for (\ref{eq:gaussian_family}) makes it possible to evaluate (\ref{eq:message_variable_to_factor}) and (\ref{eq:message_factor_to_variable}) formally even for $n\in \mathbb{R}$.
    The limit $n\to0$ is taken based on the resulting expressions.
    Application of this procedure to the factor graph in Figure \ref{fig:factor_graph} provides the Algorithm \ref{algo:rVAMP}, which we name the \emph{replicated vector approximate message passing} (rVAMP).

    At convergence, we obtain the following two-types of approximate densities for the replicated system $\tilde{p}^{(\beta)}(\bm{x})$:
    \begin{align}
        \bar{p}_1^{(\beta)}(\bm{x}) &=  \int \frac{1}{\bar{Z}_1^{(\beta)}(\xi)}\left\{ \mathbb{E}_{\Theta}\left[p_0(x;\Theta)\right]\prod_{i=1}^N \mathcal{N}\left(x_i;r_{1x,i} + \hat{Q}_{1x,i}^{-1}\sqrt{\hat{\chi}_{1x,i}}\xi_i, \hat{Q}_{1x,i}^{-1}\right) \right\}^\beta D^N\xi , \label{eq:approximate_one}\\
        \bar{p}_2^{(\beta)}(\bm{x}) &= \int \left\{\mathcal{N}\left(
            x; m_2(\xi_x, \xi_u) , \Lambda_x^{-1}
            \right)\right\}^\beta D^N\xi_xD^N\xi_u, \label{eq:approximate_two}
    \end{align}
    where 
    \begin{align}
        \bar{Z}_1^{(\beta)}(\xi) &= \int \left\{\mathbb{E}_{\Theta}\left[p_0(x;\Theta)\right]\prod_{i=1}^N \mathcal{N}\left(x_i;r_{1x,i} + \hat{Q}_{1x,i}^{-1}\sqrt{\hat{\chi}_{1x,i}}\xi_i, \hat{Q}_{1x,i}^{-1}\right)\right\}^{\beta}dx,\\
        m_2(\xi_x,\xi_u) &= \Lambda_x^{-1}\left(\Diag(\hat{Q}_{2x})r_{2x} + A^\top r_{2u} +\sqrt{\hat{\chi}_{2x}}\circ\xi_x + A^\top \left(\frac{\sqrt{\hat{\chi}_{2u}}}{\hat{Q}_{2u}}\circ\xi_u\right)\right).
    \end{align}
    $D\xi$ denotes a notation for a Gaussian measure $e^{-\xi^2/2}/\sqrt{2\pi}d\xi$.
    Additionally, we use $\Diag(x)$ for the diagonal matrix wherein diagonal elements are $(x_1,x_2,...)$, and $x\circ y$ for the Hadamard product.
    The Gaussian densities for $\xi$s and the integrand for these Gaussians correspond to the approximate densities of $p^{(\beta)}(\xi)$ and $p^{(\beta)}(f;\xi)$ in (\ref{eq:de-finetti}), respectively.
    
    The advantages of rVAMP are as follows.
    First, the approximate densities are tractable in several useful problems.
    If the prior and hyperparameter distributions are separable, then $\bar{p}_1^{(\beta)}$ is also separable: 
    $\bar{p}_1^{(\beta)}(\bm{x}) \doteq \prod_{i=1}^N\int (\bar{Z}_1^{(\beta)}(\xi_i))^{-1} \{\mathbb{E}_{\Theta_i}[p_0(x_i;\Theta_i)]\mathcal{N}(x_i ; r_{1x,i} + \sqrt{\hat{\chi}_{1x,i}} /\hat{Q}_{1x,i} \xi_i, \hat{Q}_{1x,i}^{-1})\}^{\beta} D\xi_i$, where
    $\bar{Z}_1^{(\beta)}(\xi_i) = \int \{\mathbb{E}_{\Theta_i}[p_0(x_i;\Theta_i)]\mathcal{N}(x_i ; r_{1x,i} + \sqrt{\hat{\chi}_{1x,i}} /\hat{Q}_{1x,i} \xi_i, \hat{Q}_{1x,i}^{-1})\}^{\beta}dx_i$.
    Furthermore, $\bar{p}_2^{(\beta)}$ is tractable because it is a Gaussian.
    Second, by adjusting the parameter $\beta$, the method is applicable to both Bayes inference ($\beta=1$) and MAP ($\beta\to\infty$) estimation.
    Finally, VAMP shares the same fixed point with the EC and adaptive TAP. 
    This indicates that the proposed method is expected to offer the exact results for problems of a certain class in a large system limit \cite{opper2001adaptive, opper2001tractable}.

    By construction, the two densities are constrained to have identical first and second diagonal moments \cite{rangan2017vector, fletcher2016expectation}: $\mathbb{E}_{x_i}[x_i;\bar{p}_1^{(\beta)}(x_i)]=\mathbb{E}_{x_i}[x_i;\bar{p}_2^{(\beta)}(x_i)]$,  $\mathbb{E}_{x_i}[x_i^2; \bar{p}_1^{(\beta)}(x_i)] = \mathbb{E}_{x_i}[x_i^2;\bar{p}_2^{(\beta)}(x_i)]$,  $i=1,2,...,N$.
    However, for off-diagonal moments, $\bar{p}_2^{(\beta)}$ is argued to be more precise than $\bar{p}_1^{(\beta)}$\cite{opper2005expectation, NIPS2003_2429}.
    Similarly, it is expected that for higher-order diagonal moments, $\bar{p}_1^{(\beta)}$ is more precise than $\bar{p}_2^{(\beta)}$ because it incorporates non-Gaussianity of the estimator's distribution.
    Thus, the two distributions should be used depending on the objective.

    Because the heaviest part of rVAMP is the matrix inverse and matrix-matrix product computation in line \ref{line:inverse} of Algorithm \ref{algo:rVAMP}, its computational complexity is $\mathcal{O}(N_{\rm iter}(\min(N, M)^3 + \max(N, M)^2))$ where $N_{\rm iter}$ denotes the number of iterations at convergence.
    Here the computational complexity of inverse computation is reduced to $\mathcal{O}(\min(M,N)^3)$ from $\mathcal{O}(\max(M,N)^3)$ via Woodbury formula \cite{golub1996matrix}.
    This property is quite preferable, especially in high-dimensional statistics where the number of samples $M$ is much smaller than that of the parameter $N$.
    In experiments, we empirically observed that the algorithm typically converges with $\mathcal{O}(10)$ iterations for tolerance $\delta_{\rm tol}=10^{-12}$.
    
    \begin{algorithm}[h]
    \caption{rVAMP}
    \begin{algorithmic}[1]  \label{algo:rVAMP}
    \REQUIRE{
         tolerance $\delta_{\rm tol}$, distributions $p_0(x;\Theta), \hat{p}_0(u;c)=\prod_\mu\hat{p}_0(u_\mu;c_\mu)$, $p(\Theta)$, feature matrix $A\in\mathbb{R}^{M\times N}$ whose raws are $a_1, a_2, ...,a_M$, and $p(c)=\prod_\mu {\rm Poisson}(c_\mu;M_B/M)$.
         \label{require:require}
    }
    \STATE{Initialize $r_{1x}, r_{1u}, \hat{Q}_{1x}, \hat{Q}_{1u}$.}
    \WHILE{$\delta > \delta_{\rm tol}$ \label{line:tol}}
        \STATE{// Variable $1$}
        \STATE{$\hat{x}_1=\mathbb{E}_{\Theta,\xi_{x}}\left[ \mathbb{E}_{x_1}\left[x_1 ; \left\{p_0(x_1;\Theta)\mathcal{N}(x_1;r_{1x} + \Diag(\sqrt{\hat{\chi}_{1x}}/\hat{Q}_{1x}) \circ\xi_x, \Diag(\hat{Q}_{1x}^{-1}))\right\}^\beta\right]\right]$}
        \STATE{$\hat{u}_1=\mathbb{E}_{c,\xi_u}\left[ \mathbb{E}_{u_1}\left[u_1 ; \left\{\hat{p}_0(u_{1};c)\mathcal{N}(u_1;r_{1u} + \Diag(\sqrt{\hat{\chi}_{1u}}/\hat{Q}_{1u}) \circ\xi_u, \Diag(\hat{Q}_{1u}^{-1}))\right\}^\beta\right]\right]$}
        \STATE{$v_{1x}=\mathbb{E}_{\Theta,\xi_x}\left[ \mathbb{E}_{x_1}\left[x_1 ; \left\{p_0(x_1;\Theta)\mathcal{N}(x_1;r_{1x} + \Diag(\sqrt{\hat{\chi}_{1x}}/\hat{Q}_{1x}) \circ\xi_x, \Diag(\hat{Q}_{1x}^{-1}))\right\}^\beta\right]^2 \right]-\hat{x}_1^2$}
        \STATE{$v_{1u} = \mathbb{E}_{c,\xi_u}\left[ \mathbb{E}_{u_1}\left[u_1 ; \left\{\hat{p}_0(u_{1};c)\mathcal{N}(u_1;r_{1u} + \Diag(\sqrt{\hat{\chi}_{1u}}/\hat{Q}_{1u}) \circ\xi_u, \Diag(\hat{Q}_{1u}^{-1}])\right\}^\beta\right)^2\right]-\hat{u}_1^2$}
        \STATE{$\chi_{1x} = \hat{Q}_{1x}^{-1}\circ \partial_{r_{1x}} \hat{x}_{1}, \quad \chi_{1u} = \hat{Q}_{1u}^{-1}\circ \partial_{r_{1u}}\hat{u}_1$}
        \STATE{$\eta_{1x,1}=\chi_{1x}^{-1}, \, \eta_{1x,2} = v_{1x}\ \circ (\chi_{1x}^{-2}), \quad \eta_{1u,1}=\chi_{1u}^{-1}, \, \eta_{1u,2}=v_{1u}\circ (\chi_{1u}^{-2})$}
        
        \STATE{$\hat{Q}_{2x}=\eta_{1x,1} -\hat{Q}_{1x} , \; \hat{\chi}_{2x}=\eta_{1x,2} - \hat{\chi}_{1x}, \; r_{2x}=\hat{Q}_{2x}^{-1}\circ(\eta_{1x,1}\circ \hat{x}_1 - \hat{Q}_{1x}\circ r_{1x})$}
        \STATE{$\hat{Q}_{2u}=\eta_{1u,1} -\hat{Q}_{1u} , \; \hat{\chi}_{2u}=\eta_{1u,2} - \hat{\chi}_{1u}, \; r_{2u}=\hat{Q}_{2u}^{-1}\circ(\eta_{1u,1}\circ \hat{x}_1 - \hat{Q}_{1u}\circ r_{1u})$}
        \STATE{\vspace{-2pt}}
        
        \STATE{// Variable 2}
        \STATE{$\Lambda_x = \Diag(\hat{Q}_{2x}) + A^\top \Diag(\hat{Q}_{2u}^{-1})A, \quad \Lambda_u = \Diag(\hat{Q}_{2u}) + A \Diag(\hat{Q}_{2x}^{-1})A^\top$}
        \STATE{$\hat{x}_2=\Lambda_x^{-1}(A^\top r_{2u}+\hat{Q}_{2x}\circ r_{2x}), \quad \hat{u}_2=\Lambda_u^{-1}(-A r_{2x} + \hat{Q}_{2u}\circ r_{2u} )$} \label{line:inverse}
        \STATE{$v_{2x} = (\Lambda_x\circ\Lambda_x) \hat{\chi}_{2x} + [(\Lambda_x A^\top)\circ(\Lambda_x A^\top)](\hat{\chi}_{2u}\circ\hat{Q}_{2u}^{-2})$}

        \STATE{$v_{2u} = (\Lambda_u\circ\Lambda_u) \hat{\chi}_{2u} + [(\Lambda_u A )\circ(\Lambda_u A )](\hat{\chi}_{2x}\circ\hat{Q}_{2x}^{-2})$}
        \STATE{$\chi_{2x} = \hat{Q}_{2x}^{-1}\circ \partial_{r_{2x}} \hat{x}_{2}, \quad \chi_{2u} = \hat{Q}_{2u}^{-1}\circ \partial_{r_{2u}}\hat{u}_2$}
        \STATE{$\eta_{2x,1}=\chi_{2x}^{-1}, \, \eta_{2x,2} = v_{2x}\ \circ (\chi_{2x}^{-2}), \quad \eta_{2u,1}=\chi_{2u}^{-1}, \, \eta_{2u,2}=v_{2u}\circ (\chi_{2u}^{-2})$}
        %
        \STATE{$\hat{Q}_{1x}=\eta_{2x,1} -\hat{Q}_{2x} , \; \hat{\chi}_{2x}=\eta_{2x,2} - \hat{\chi}_{2x}, \; r_{2x}=\hat{Q}_{1x}^{-1}\circ(\eta_{2x,1}\circ \hat{x}_2 - \hat{Q}_{2x}\circ r_{2x})$}
        \STATE{$\hat{Q}_{1u}=\eta_{2u,1} -\hat{Q}_{2u} , \; \hat{\chi}_{2u}=\eta_{2u,2} - \hat{\chi}_{2u}, \; r_{2u}=\hat{Q}_{1u}^{-1}\circ(\eta_{2u,1}\circ \hat{x}_2 - \hat{Q}_{2u}\circ r_{2u})$}
        \STATE{}
        \STATE{$\delta = \max(\|\hat{x}_1 - \hat{x}_2\|_2/\sqrt{N}, \|v_{1x} - v_{2x}\|_2/\sqrt{N}) \quad $ // convergence criterion}
        \label{line:delta}
        
    \ENDWHILE
    \STATE{Return $r_{1x}, \hat{Q}_{1x}, \hat{\chi}_{1x}, r_{2x}, \hat{Q}_{2x}, \hat{Q}_{2u}$, $r_{1u}, \hat{Q}_{1u}, \hat{\chi}_{1u}, r_{2u}, \hat{Q}_{2u}, \hat{Q}_{2u}$, $\Lambda_x, \Lambda_u$.}
    \end{algorithmic}
    \end{algorithm}

\section{Experiment}
    \label{sec:experiment}
    We apply rVAMP to stability selection (SS) \cite{meinshausen2010stability}, which is a resampling-based variable selection technique.
    The purpose of the experiment involves evaluating the computational efficiency and approximation accuracy of rVAMP for synthetic and real-world data.
    Experiments are conducted on a single processor of a 2.6-GHz Intel Core i7.
    The codes used in experiments are on \cite{experimentcode}.    
    
    \subsection{Stability selection}
    We consider SS for variable selection problems in sparse linear regression with $\ell_1$ penalty that is also termed as LASSO \cite{tibshirani1996regression}.
    In linear regression, each data point $z_\mu = (a_\mu, y_\mu)$ consists of a feature $a_\mu\in\mathbb{R}^N$ and output $y_\mu\in\mathbb{R}$ that is given by $y_\mu = a_\mu^\top x_0 + w_\mu$, $w_\mu \sim \mathcal{N}(w_\mu;0, \sigma^2)$, $\mu=1,2,...,M$.
    With the notation $S(x) = {\rm supp}(x_0)\subseteq\{1,2,...,N\}$, the goal of the variable selection is to determine $S(x_0)$ from $D=\{(a_\mu, y_\mu)\}_{\mu=1}^M$.
    To this aim, LASSO seeks an estimator as
    \begin{equation}
        \hat{x}(D, \lambda) = \arg\min_{x}\left[\frac{1}{2}\sum_{\mu=1}^M(y_\mu - a_\mu^\top x)^2 + \sum_{i=1}^N\lambda_i|x_i|\right], \quad  \lambda_i >0, i=1,2,...,N. \label{eq:lasso}
    \end{equation}
    The $\ell_1$ regularization allows LASSO to select variables by shrinking certain estimated parameters exactly to 0.
    However, even for $\{\lambda_i\}$ chosen optimally by cross validation, generally $S(\hat{x})$ contains false positive elements; i.e. there are elements of $S(\hat{x})$ that are not included in $S(x_0)$.
    
    SS is a method to suppress such disadvantage of variable selection ability of LASSO.
    The basic concept of SS is to consider the bootstrapped distribution of $D$ and hyperparameter distribution $P(\lambda)$ and to calculate the probability $\Pi_i={\rm Prob}[{\hat{x}_i\neq0}]$, $i=1,2,..,N$.
    \cite{meinshausen2010stability} showed that for a proper choice of bootstrapped sample size and hyperparameter distribution $P(\lambda)$, we can reduce the amount of false positive elements by focusing on $\Pi$ as opposed to $\hat{x}(D, \lambda)$. 
    The conventional choice of the size of bootstrapped sample $M_B$ is $M/2$ and that of $P(\lambda)$ is $P(\lambda) = \prod_i P(\lambda_i)$, $P(\lambda_i)=\{\delta(\lambda_i - 2\lambda) + \delta(\lambda_i - \lambda)\}/2$, $\lambda>0$.
    The major disadvantage of SS is the computational cost because the distribution of the LASSO estimator is typically obtained by numerically solving the optimization problem (\ref{eq:lasso}) for each resampled $(D, \lambda)$.
    
    In the notation described in Section \ref{sec:resampling problem}-\ref{sec:rVAMP}, the above setting corresponds to $l(z_\mu; f) = (y_\mu - a_\mu^\top x)^2/2$, $M_B=M/2$, $p_0(x;\Theta) \doteq \prod_i \exp(-\lambda_i|x_i|)$ and $\beta\to\infty$, respectively.
    
    \subsection{Synthetic dataset}
    The first experiment considers the linear regression on a synthetic dataset.
    In the experiment, we generate true parameter $x_0\in\mathbb{R}^N$ based on Bernoulli-Gauss model $x_{0,i}\sim_{\rm i.i.d.} \rho \mathcal{N}(x_{0,i};0,1) + (1-\rho)\delta(x_{0,i})$, $i=1,2,...,N$ and set the size of parameter $N=4096$, size of measurements $M=\alpha N, \alpha=0.15, 0.1, 0.09, 0.075, 0.05$, sparsity $\rho = 0.05$, and standard deviation of measurement noise $\sigma=0.1$.
    The features $\{a_\mu\}$ are created as random $M$ row selection from a discrete cosine transform matrix (random DCT).
    Both features and outputs are centered and normalized as $\sum_\mu a_{\mu i}=0, \sum_\mu a_{\mu i}^2 = 1, i=1,2,...,N$, $\sum_\mu y_\mu=0$.
    Regularization strength $\lambda$ is selected via a 10-fold cross validation.
    We set the tolerance $\delta_{\rm tol}$ in line \ref{line:tol} of Algorithm \ref{algo:rVAMP} at $10^{-12}$.

    Figure \ref{fig:time evolution} shows the time evolution of the difference $\delta$ of the two approximate densities (\ref{eq:approximate_one}) and (\ref{eq:approximate_two}) defined in line \ref{line:delta} of Algorithm \ref{algo:rVAMP}.
    The difference is plotted for different measurement ratios $\alpha=M/N$.
    In all cases, the differences exhibit plain exponential decay relative to the iteration step $t$.
    This demonstrates a fast convergence of the rVAMP.
    
   To check the approximation accuracy for the synthetic data, we compare the rVAMP estimates of mean $\mathbb{E}[\hat{x}_i]$, variance ${\rm Var}[\hat{x}_i]$, and $\Pi_i = {\rm Prob}[\hat{x}_i\neq0]$, $i=1,2,...,N$ with those obtained via naive resampling and re-estimation.
    With respect to the naive estimate, we create $100,000$ samples of bootstrapped data $(D, \lambda)$ and use Matlab implementation of Glmnet \cite{glmnetmatlab} to obtain LASSO estimators.
    With respect to the rVAMP estimate, we use approximate distribution (\ref{eq:approximate_one}) to accurately incorporate non-Gaussianity of statistics.
    In this case, the density (\ref{eq:approximate_one}) is factorized.
    The upper panels of Figure \ref{fig:accuracy} show the comparison of rVAMP estimate with the naive estimate.
    All the statistics exhibit almost complete agreement as expected from the recent analysis of EC approximate inference and adaptive TAP for a linear model with random DCT observation matrix \cite{ccakmak2018expectation, takahashi2018statistical}.
    
    \subsection{Real world dataset: riboflavin dataset}
    The riboflavin dataset \cite{buhlmann2014high} is used as real-world data.
    This is a genomic dataset concerning riboflavin (vitamin B2) production rate and is commonly used as a test-bed in high-dimensional statistics.
    The data consists of $M=71$ pairs of real-valued output and $N=4088$ dimensional feature that corresponds to gene expression.
    Pre-processing which includes hyperparameter setting is identical to that of synthetic data experiment.
    
\begin{figure}[h]
      \centering
      \includegraphics[width= \columnwidth]{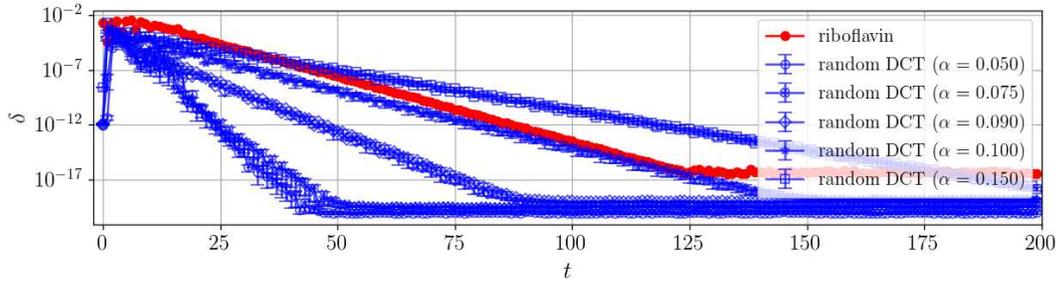}
      \caption{Time evolution of the difference $\delta$ for both synthetic and real-world dataset. Blue filled symbols denote the synthetic data experiment, and the red open circles denote the real-world data experiment. For both datasets, the difference decays exponentially.
      For random DCT data, the average is taken over $10$ samples. We evaluate the error bar as one standard error.
      }
      \label{fig:time evolution}
    \end{figure}

\begin{figure}[h]
      \centering
      \includegraphics[width= \columnwidth]{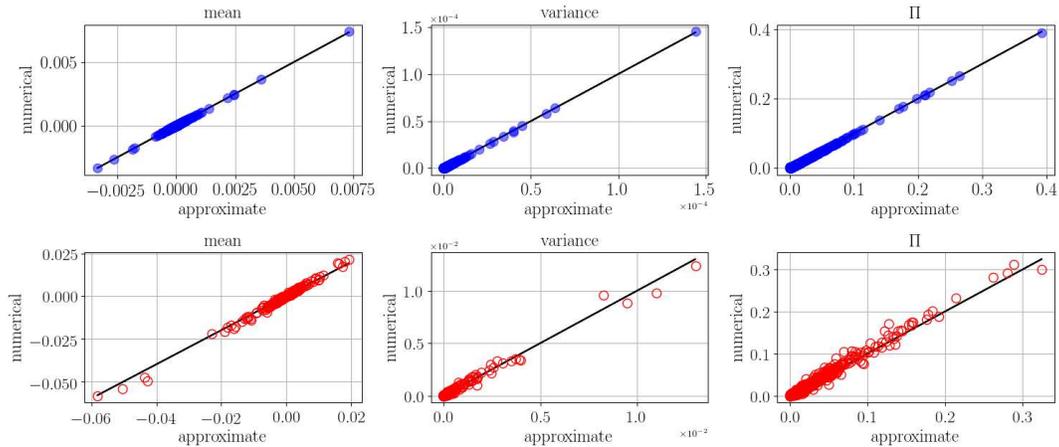}
      \caption{Naive numerical resampling estimate of $\mathbb{E}[\hat{x}_i]$, variance ${\rm Var}[\hat{x}_i]$ and $\Pi_i = {\rm Prob}[\hat{x}_i]$, $i=1,2,...,N$ are compared with that of the rVAMP estimate. The upper panels denote the synthetic data result and the lower panels denote the real-world data result.
      With respect to synthetic data, almost complete agreement is observed. With respect to real-world data, good agreement is observed. 
      }
      \label{fig:accuracy}
    \end{figure}

    Figure \ref{fig:time evolution} and the lower panels of Figure \ref{fig:accuracy} exhibit the same time evolution and comparison as that of the synthetic data experiment.
    In a manner similar to the synthetic data case, the time evolution of the difference $\delta$ exhibits exponential decay and
    the comparison of the three statistics exhibits good agreement.
    The results also demonstrate the usefulness of rVAMP for the real-world dataset.
    
\section{Conclusion}
    In this study, we developed an efficient approximate inference algorithm for resampling average of estimators.
    The key idea involves constructing the VAMP algorithm on the replicated system using the replica method and variable augmentation by Fourier transformation.
    Application to a resampling-based variable selection method called stability selection in synthetic and real-world datasets indicated that the convergence criterion exhibits exponential decay in the iteration step and the algorithm offers excellent approximation accuracy.
    Promising future work includes an extension of the current scheme to the stochastic algorithm that naturally scales to a larger dataset, analysis of convergence dynamics based on so-called \emph{state evolution} \cite{donoho2009message, bayati2011dynamics, rangan2017vector} and theoretical analysis of resampling methods based on the replica theory of statistical physics \cite{mezard1987spin, nishimori2001statistical, dotsenko2005introduction,zamponi2010mean}.
    
\bibliographystyle{amsalpha}
\bibliography{references}

\end{document}